\documentclass{article}

\usepackage[final,nonatbib]{nips_2016}

\usepackage[utf8]{inputenc} 
\usepackage[T1]{fontenc}    
\usepackage{hyperref}       
\usepackage{url}            
\usepackage{booktabs}       
\usepackage{amsfonts}       
\usepackage{nicefrac}       
\usepackage{microtype}      
\usepackage{amsmath}
\usepackage{color}
\usepackage{graphicx}

\usepackage{floatrow}
\newfloatcommand{capbtabbox}{table}[][\FBwidth]

\newcommand{\fig}[1]{Fig.~\ref{fig:#1}}
\newcommand{\tab}[1]{Table~\ref{tab:#1}}
\newcommand{\secc}[1]{Section~\ref{sec:#1}}
\newcommand{\append}[1]{Appendix~\ref{app:#1}}
\def\etal{{\textit{et~al.~}}}
\newcommand{\res}[2]{#1{\scriptsize $\pm$ #2} }

\newcommand {\R}{\mathbb {R}}

\title{Learning Multiagent Communication 
         \\with Backpropagation}

\author{
Sainbayar Sukhbaatar\\
Dept. of Computer Science\\
Courant Institute, New York University\\
\texttt{sainbar@cs.nyu.edu}
\And
Arthur Szlam\\
Facebook AI Research \\
New York\\
\texttt{aszlam@fb.com}
\And
Rob Fergus \\
Facebook AI Research \\
New York\\
\texttt{robfergus@fb.com}
}

\begin{document}

\maketitle
\vspace{-2mm}
\begin{abstract}
\vspace{-2mm}
  Many tasks in AI require the collaboration of multiple
  agents. Typically, the communication protocol between agents is
  manually specified and not altered during training. In this
  paper we explore a simple neural model, called CommNet, that uses continuous
  communication for fully cooperative tasks. The model consists of
  multiple agents and the communication between them is learned
  alongside their policy. We apply this model to a diverse set of
  tasks, demonstrating the ability of the agents to learn to
  communicate amongst themselves, yielding improved performance over
  non-communicative agents and baselines. In some cases, it is
  possible to interpret the language devised by the agents, revealing
  simple but effective strategies for solving the task at hand.
\end{abstract}

\vspace{-4mm} 
\section{Introduction}
\vspace{-2mm} 
Communication is a fundamental aspect of intelligence, enabling agents
to behave as a group, rather than a collection of individuals. It is
vital for performing complex tasks in real-world environments where
each actor has limited capabilities and/or visibility of the world. Practical examples
include  elevator
control \cite{Crites1998} and sensor networks \cite{Fox00}; communication is also important
for success in robot soccer \cite{Stone98}.
In any partially observed environment, the communication between
agents is vital to coordinate the behavior of each individual. While
the model controlling each agent is typically learned via
reinforcement learning \cite{busoniu2008comprehensive,Sutton98}, the
specification and format of the communication is usually
pre-determined. For example, in robot soccer, the bots are designed to
communicate at each time step their position and proximity to the
ball.

In this work, we propose a model where cooperating agents learn to
communicate amongst themselves before taking actions. Each agent is controlled by a deep
feed-forward network, which additionally has access to a communication
channel carrying a continuous vector. Through this
channel, they receive the summed transmissions of other
agents. However, what each agent transmits on the channel is not
specified a-priori, being learned instead.  Because the communication
is continuous, the model can be trained via back-propagation, and thus
can be combined with standard single agent RL algorithms or supervised learning.  The model
is simple and versatile. This allows
it to be applied to a wide range of problems involving partial
visibility of the environment, where the agents learn a task-specific
communication that aids performance. 
In addition, the model allows dynamic variation at run
time in both the number and type of agents, which is important in applications 
such as communication between moving cars.

We consider the setting where we have $J$ agents, all cooperating to
maximize reward $R$ in some environment.  We make the simplifying
assumption of full cooperation between agents, thus each agent receives $R$ independent of their
contribution. In this setting, there is no difference between each
agent having its own controller, or viewing them as pieces of a larger
model controlling all agents. Taking the latter perspective, our controller
is a large feed-forward neural network that maps inputs for all agents
to their actions, each agent occupying a subset of units. A specific
connectivity structure between layers (a) instantiates the broadcast
communication channel between agents and (b) propagates the agent
state.

We explore this model on a range of tasks. In some,
supervision is provided for each action while for others it is given
sporadically. In the former case, the controller for each agent is trained
by backpropagating the error signal through the connectivity
structure of the model, enabling the agents to learn how to
communicate amongst themselves to maximize the objective. In the
latter case, reinforcement learning must be used as an additional outer loop to
provide a training signal at each time step (see \append{reinforce} for
details).

\vspace{-2.5mm}
\section{Communication Model}
\vspace{-2mm}
We now describe the model used to compute the distribution over
actions $p(\mathbf{a}(t)|\mathbf{s}(t),\theta)$ at a
given time $t$ (omitting the time index for brevity). 
Let $s_j$ be the $j$th agent's view of the state of the
environment.  The input to the controller is the concatenation of all
state-views $\mathbf{s} = \{s_1,..., s_J\}$,
and the controller $\Phi$ is a mapping
$\mathbf{a} = \Phi(\mathbf{s})$, where the output $\mathbf{a}$ is a
concatenation of discrete actions $\mathbf{a} = \{a_1, ... ,a_J\}$ for each
agent. Note that this single controller $\Phi$ encompasses the
individual controllers for each agents, as well as the communication
between agents.

\vspace{-3mm}
\subsection{Controller Structure}
\vspace{-1.5mm} 
We now detail our architecture for $\Phi$ that is built from modules $f^i$, which take the form of multilayer
neural networks. Here $i\in \{0,..,K\}$, where $K$ is the number of
communication steps in the network.  

Each $f^i$ takes two input vectors for each agent $j$: the hidden state $h_j^i$ and
the communication $c_j^i$, and outputs a vector $h_j^{i+1}$.  The
main body of the model then takes as input the concatenated vectors
$\mathbf{h}^0 = [h^0_1, h^0_2, ..., h^0_J]$, and computes:
\begin{eqnarray} 
h^{i+1}_j &=& f^i(h^{i}_j, c^{i}_j) \label{eq:streamlayer} \\
c^{i+1}_j &=& \frac{1}{J-1} \sum_{j'\neq j} h^{i+1}_{j'}.\label{eq:mean_communication} 
\end{eqnarray}
In the case that $f^i$ is a single linear layer followed by a
non-linearity $\sigma$, we have:
$h^{i+1}_j = \sigma(H^ih^{i}_j+ C^ic^{i}_j)$ and the model can be
viewed as a feedforward network with layers $\mathbf{h}^{i+1} = \sigma(T^{i}\mathbf{h}^{i})$ where
$\mathbf{h}^i$ is the concatenation of all $h^i_j$ and $T^i$ takes the
block form (where $\bar{C}^i = C^i/(J-1)$):

{\small
\[T^i = \begin{pmatrix}
	H^i & \bar{C}^i & \bar{C}^i & ... & \bar{C}^i \\
 	\bar{C}^i & H^i & \bar{C}^i &... & \bar{C}^i \\	
	\bar{C}^i & \bar{C}^i & H^i & ... & \bar{C}^i \\
\vdots  & \vdots &\vdots & \ddots & \vdots \\
	\bar{C}^i & \bar{C}^i & \bar{C}^i &... & H^i 
	\end{pmatrix}
,	\]}

        A key point is that $T$ is {\it dynamically sized} since 
        the number of agents may vary. This motivates the the
        normalizing factor $J-1$ in equation
        \eqref{eq:mean_communication}, which rescales the
        communication vector by the number of communicating
        agents. Note also that $T^i$ is
        permutation invariant, thus the order of the agents does not matter.

\begin{figure}[h!]
\centering
\includegraphics[width=0.9\linewidth]{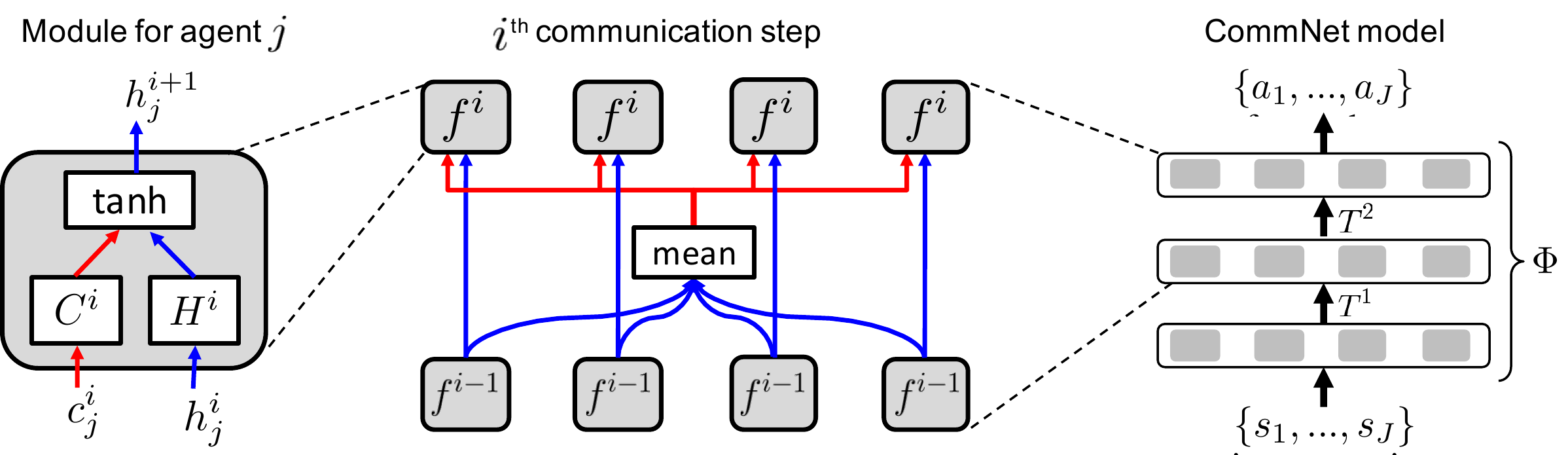}
\caption{An overview of our CommNet model. Left: view of module
  $f^i$ for a single agent $j$. Note that the parameters are shared
  across all agents. Middle: a single communication step, where each
  agents modules propagate their internal state $h$, as well as
  broadcasting a communication vector $c$ on a common channel (shown
  in red). Right: full model $\Phi$, showing input states $s$ for each agent,
two communication steps and the output actions for each agent.}
\label{fig:model1}
\end{figure}

        At the first layer of the model an encoder function $h^{0}_j =r(s_j$)
        is used. This takes as input state-view $s_j$ and outputs
        feature vector $h^{0}_j$ (in $\R^{d_0}$ for some $d_0$). The
        form of the encoder is problem dependent, but for most of our
        tasks it is a single layer neural network. Unless otherwise noted, $c^{0}_j = 0 $ for all $j$.      
        At the output of the model, a decoder function $q(h^K_j)$ is
        used to output a distribution over the space of
        actions. $q(.)$ takes the form of a single layer network,
        followed by a softmax. To produce a discrete action, we sample
        from this distribution: $a_j \sim q(h^K_j)$.

        Thus the entire model (shown in \fig{model1}), which we call a Communication Neural
        Net (CommNet), (i) takes the state-view of all agents $\mathbf{s}$,
        passes it through the encoder $\mathbf{h}^0 = r(\mathbf{s})$, (ii)
        iterates $\mathbf{h}$ and $\mathbf{c}$ in equations
        \eqref{eq:streamlayer} and \eqref{eq:mean_communication} to
        obtain $\mathbf{h}^K$, (iii) samples actions $\mathbf{a}$ for
        all agents, according to $q(\mathbf{h}^K)$.

\vspace{-2mm}
\subsection{Model Extensions}
\vspace{-2mm}
\label{sec:extra}
\noindent {\bf Local Connectivity:}  An alternative to the broadcast
framework described above is to allow agents to communicate to others
within a certain range. Let $N(j)$ be the set of agents present within
communication range of agent $j$. Then \eqref{eq:mean_communication} becomes:
\begin{equation} 
c^{i+1}_j = \frac{1}{|N(j)|} \sum_{j'\in N(j)}  h^{i+1}_{j'}.
\label{eq:cp1}
\end{equation}
As the agents move, enter and exit the environment, $N(j)$ will
change over time.  
In this setting, our model has a natural interpretation as a dynamic
graph, with $N(j)$ being the set of vertices connected to vertex $j$
at the current time. The edges within the graph represent the
 communication channel between agents, with \eqref{eq:cp1} being
 equivalent to belief propagation \cite{pearl1982reverend}.
 Furthermore, the use
 of multi-layer nets at each vertex makes our model similar to an 
 instantiation of the GGSNN work of Li \etal \cite{GGSNN}. 

\noindent {\bf Skip Connections:} For some tasks, it is useful to have
the input encoding $h^0_j$ present as an input for communication
steps beyond the first layer. Thus for agent $j$ at step $i$, we have: 
\begin{equation} h^{i+1}_j = f^i(h^{i}_j, c^{i}_j,h^0_j ) \label{eq:skip}.\end{equation}

\noindent {\bf Temporal Recurrence:}  We also explore having the network be
a recurrent neural network (RNN). This is achieved by simply replacing
the communication step $i$ in Eqn.~\eqref{eq:streamlayer} and
\eqref{eq:mean_communication} by a time step $t$, and using the same
module $f^t$ for all $t$. At every time step, actions will be sampled
from $q(h^t_j)$. Note that agents can leave or join the swarm at any time step.
If $f^t$ is a single layer network, we obtain plain RNNs that
communicate with each other. In later experiments, we also use an LSTM
as an $f^t$ module.

\vspace{-2mm}
\section{Related Work}
\vspace{-2mm}

Our model combines a deep network with reinforcement learning \cite{Guo14,Mnih15,levine15}. Several
recent works have applied these methods to multi-agent domains, such
as Go \cite{Maddison15,Silver16} and Atari games
\cite{Tampuu15}, but they assume full visibility of the environment and
lack communication. 
There is a rich literature on multi-agent reinforcement learning (MARL)
\cite{busoniu2008comprehensive}, particularly in the robotics domain
\cite{Matari97,Stone98,Fox00,Olfati07,Cao13}. Amongst fully cooperative algorithms, many approaches
\cite{Lauer00,littman2001value,wang2002reinforcement} avoid the need
for communication by making strong assumptions about visibility of
other agents and the environment. Others use communication, but with
a pre-determined protocol \cite{Tan93,Melo11,Zhang13,Maravall13}.

A few notable approaches involve learning to communicate between
agents under partial visibility: Kasai \etal \cite{Kasai08} and Varshavskaya \etal
\cite{Varshavskaya2009}, both use distributed tabular-RL
approaches for simulated tasks. Giles \& Jim \cite{Giles02} use an
evolutionary algorithm, rather than reinforcement learning. Guestrin \etal \cite{Guestrin01} use
a single large MDP to control a collection of agents, via a factored
message passing framework where the messages are learned. 
In contrast to these approaches, our model uses a deep network for both agent control and
communication.

From a MARL perspective, the closest approach to ours is the concurrent work
of Foerster \etal \cite{Foerster16}. This also uses a deep
reinforcement learning in multi-agent partially observable tasks,
specifically two riddle problems (similar in spirit to our {\it levers} task) which necessitate multi-agent
communication. Like our approach, the communication is learned rather
than being pre-determined. However, the agents communicate in
a discrete manner through their actions. This contrasts with our model
where multiple continuous communication cycles are used at each time
step to decide the actions of all agents. Furthermore, our approach is
amenable to dynamic variation in the number of agents.

The Neural GPU \cite{NeuralGPU} has similarities to our model but
differs in that a 1-D ordering on the input is assumed and it 
employs convolution, as opposed to the global pooling in our approach
(thus permitting unstructured inputs).
Our model can be regarded as an instantiation of the GNN construction of Scarselli
\etal \cite{GNN}, as expanded on by Li \etal \cite{GGSNN}.  In
particular, in \cite{GNN}, the output of the model is the fixed point
of iterating equations \eqref{eq:cp1} and \eqref{eq:streamlayer} to
convergence, using recurrent models. In \cite{GGSNN}, these recurrence
equations are unrolled a fixed number of steps and the model trained
via backprop through time. In this work, we do not require the model
to be recurrent, neither do we aim to reach steady state. Additionally, we regard
Eqn.~\eqref{eq:cp1} as a pooling operation, conceptually making our
model a single feed-forward network with local connections.

\vspace{-2mm}
\section{Experiments}
\vspace{-2mm}

\subsection{Baselines}
\vspace{-2mm}
We describe three baselines models for $\Phi$ to compare against our model.

\textbf{Independent controller:}
A simple baseline is where agents are controlled independently 
without any communication between them.
We can write $\Phi$ as $\mathbf{a} = \{\phi(s_1),...,\phi(s_J)\}$,
where $\phi$ is a per-agent controller applied independently. 
The advantages of this communication-free model is modularity and 
flexibility\footnote{Assuming $s_j$ includes the identity of agent $j$.}.
Thus it can deal well with agents joining and leaving the group, 
but it is not able to coordinate agents' actions.

\textbf{Fully-connected:}
Another obvious choice is to make $\Phi$ a fully-connected multi-layer neural
network, that takes concatenation of $h_j^0$ as an input
and outputs actions $\{a_1,...,a_J\}$ using multiple output softmax heads. 
It is equivalent to allowing $T$ to be an arbitrary matrix with fixed size.
This model would allow agents to communicate with each other and share views of
the environment. Unlike our model, however, it is not modular, inflexible with
respect to the composition and number of agents it controls, and
even the order of the agents must be fixed.

\textbf{Discrete communication:}
An alternate way for agents to communicate is via discrete symbols,
with the meaning of these symbols being learned during training. 
Since $\Phi$ now contains discrete operations and is not differentiable, reinforcement
learning is used to train in this setting. However, unlike actions in the 
environment, an agent has to output a discrete symbol at every
communication step. But if these are viewed as \emph{internal} time steps of the
agent, then the communication output can be treated as an action of the
agent at a given (internal) time step and we can directly employ 
policy gradient \cite{Williams92simplestatistical}.

At communication step $i$, agent $j$ will output the index $w^i_j$
corresponding to a particular symbol, sampled according to: 
\begin{equation} w^i_j \sim \text{Softmax}(D h^{i}_j) \end{equation}
where matrix $D$ is the model parameter. Let $\hat{w}$ be a 1-hot binary
vector representation of $w$. In our broadcast framework, at the next step the agent
receives a bag of vectors from all the other agents (where $\wedge$ is the element-wise OR operation):
\begin{equation} c^{i+1}_j = \bigwedge_{j'\neq j} \hat{w}^i_{j'}   \label{eqn:dcom} \end{equation}

\vspace{-2mm}
\subsection{Simple Demonstration with a Lever Pulling Task}
\vspace{-2mm}
We start with a very simple game that requires the agents to
communicate in order to win.  This consists of $m$ levers and a pool
of $N$ agents. At each round, $m$ agents are drawn at random from the total pool of $N$ agents
and they
must each choose a lever to pull, simultaneously with the other $m-1$
agents, after which the round ends. The goal is for each of them to pull a {\em different}
lever. Correspondingly, all agents receive reward proportional to the
number of distinct levers pulled. Each agent can see its own identity,
and nothing else, thus $s_j = j$. 

We implement the game
with $m=5$ and $N=500$.  We use a CommNet with two communication
steps ($K=2$) and skip connections from \eqref{eq:skip}.  The encoder $r$ is a
lookup-table with $N$ entries of $128$D.  Each $f^i$ is a
two layer neural net with ReLU non-linearities that takes in the
concatenation of $(h^i,c^i,h^0)$, and outputs a $128$D
vector.  The decoder is a linear layer plus softmax, producing a
distribution over the $m$ levers, from which we sample to determine the
lever to be pulled. We compare it against the independent controller, 
which has the same architecture as our model except that communication $c$ is zeroed.
The results are shown in Table \ref{tab:levers}. The metric is the
number of distinct levers pulled divided by $m=5$, averaged over $500$
trials, after seeing $50000$ batches of size $64$ during training. We
explore both reinforcement (see \append{reinforce}) and direct supervision
(using the solution given by sorting the agent IDs, and having each agent pull the lever according to its relative order in the current $m$ agents). In both
cases, the
CommNet performs significantly better than the independent controller. 
See \append{lever} for an analysis of a trained model.

\begin{table}[h]
\small
\center
\begin{tabular}{|l|c|c|} 
\hline
                                 & \multicolumn{2}{|c|}{Training method}         \\
Model $\Phi$       & Supervised & Reinforcement \\ \hline
Independent  &  0.59 & 0.59 \\ \hline
CommNet      & \textbf{0.99} &  \textbf{0.94} \\  \hline
\end{tabular}
\caption{
Results of lever game (\#distinct levers
  pulled)/(\#levers) for our CommNet and independent controller models,
  using two different training approaches. Allowing the agents to
  communicate enables them to succeed at the task. 
\vspace{-2mm}
  }
\label{tab:levers}
\end{table}

\subsection{Multi-turn Games}
\vspace{-2mm}
In this section, we consider two multi-agent tasks using the MazeBase
environment \cite{Mazebase} that use reward as their training
signal. The first task is to control cars passing through a traffic
junction to maximize the flow while minimizing collisions. The second task
is to control multiple agents in combat against enemy bots. 

We experimented with several module types. With a feedforward MLP, the
module $f^i$ is a single layer network and $K=2$ communication steps
are used. For an RNN module, we also used a single layer network for
$f^t$, but shared parameters across time steps. Finally, we used
an LSTM for $f^t$. In all modules, the hidden layer size is set to
50. MLP modules use skip-connections. Both tasks are trained for
300 epochs, each epoch being 100 weight updates with
RMSProp~\cite{Tieleman2012} on mini-batch of 288 game episodes
(distributed over multiple CPU cores). In total, the models experience
$\sim$8.6M episodes during training.  We repeat all experiments 5
times with different random initializations, and report mean value
along with standard deviation. The training time varies from a few hours
to a few days depending on task and module type.

\subsubsection{Traffic Junction}
\begin{figure}[h!]
\centering
\includegraphics[width=0.3\linewidth]{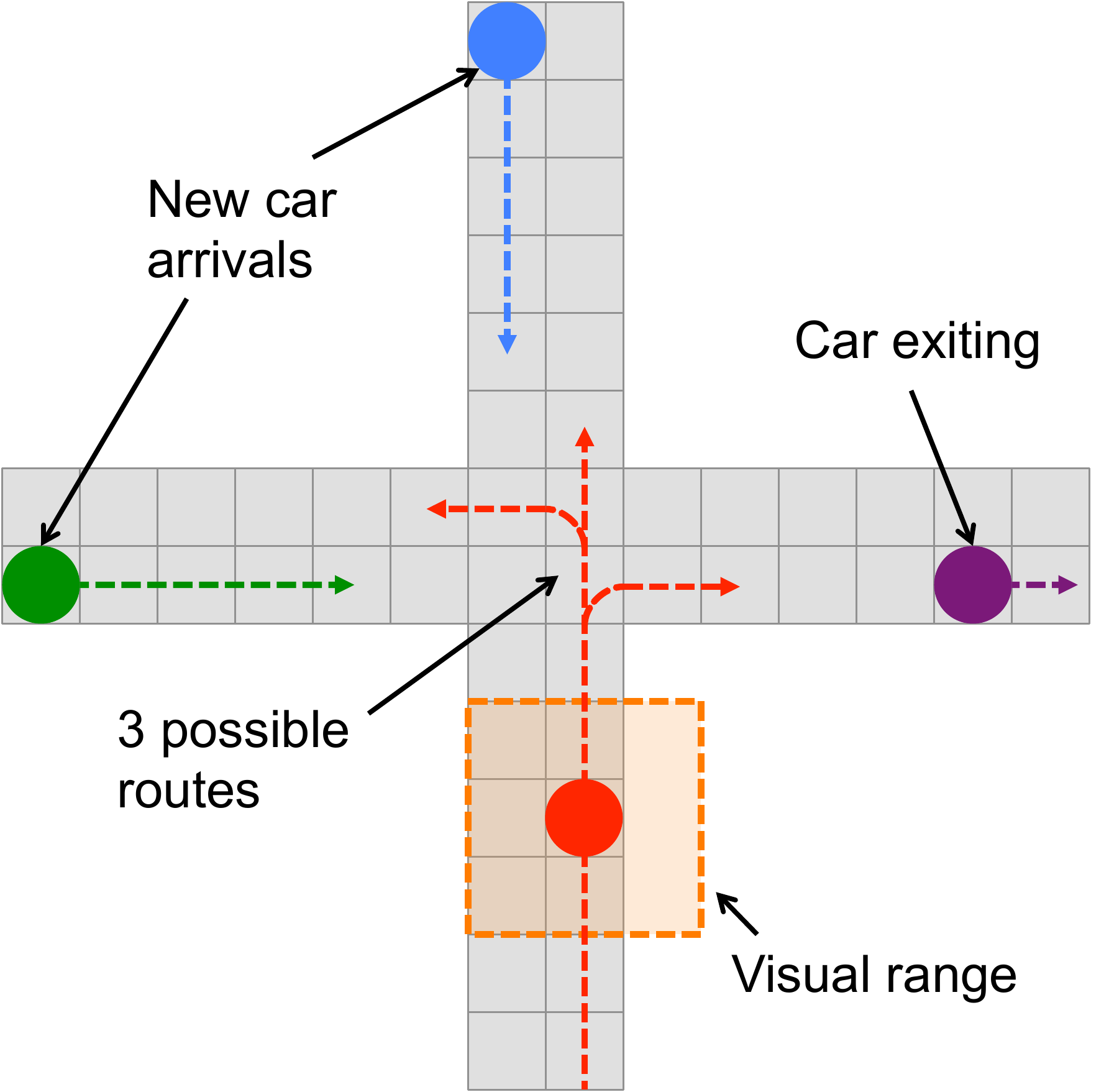}
\hspace{4mm}
\includegraphics[width=0.3\linewidth]{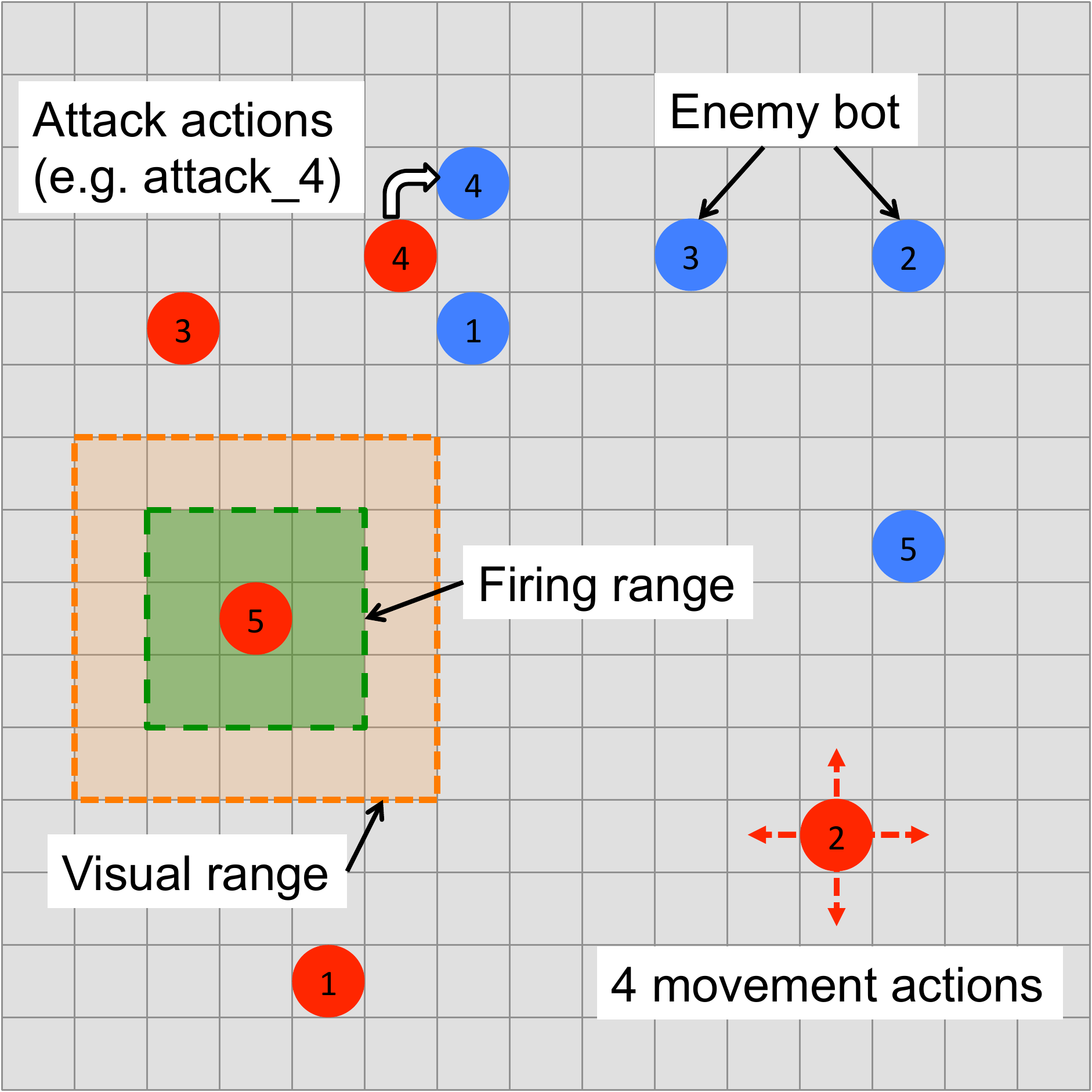}
\hspace{4mm}
\includegraphics[width=0.3\linewidth]{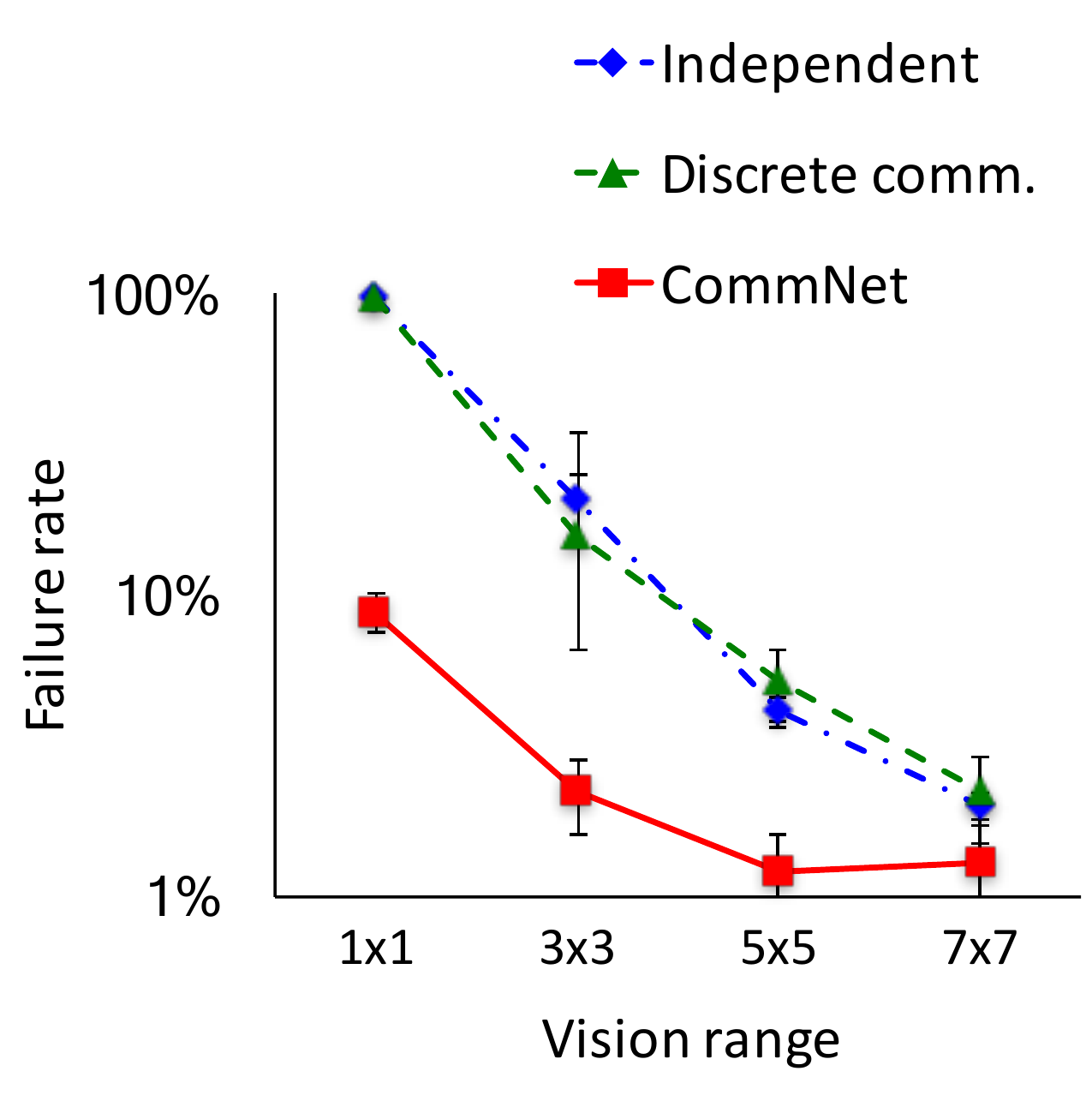}
\caption{Left: Traffic junction task where agent-controlled cars (colored circles)
  have to pass the through the junction without colliding. 
  Middle: The combat
  task, where model controlled agents (red circles) fight against
  enemy bots (blue circles). In both tasks each agent has limited
  visibility (orange region), thus is not able to see the location of all other agents.
  Right: 
  As visibility in the environment decreases, the importance of communication grows
  in the traffic junction task. }
\label{fig:junction}
\end{figure}

\vspace{-2mm}
This consists of a 4-way junction on a $14\times14$
grid as shown in \fig{junction}(left). At each time step, new cars
enter the grid with
probability $p_\text{arrive}$ from each of the four
directions. However, the total number of cars at
any given time is limited to $N_\text{max}=10$. Each car occupies a
single cell at any given time and is randomly assigned to one of three
possible routes (keeping to the right-hand side of the
road). At every time step, a car has two possible
actions: \emph{gas} which advances it by one cell on its
route or \emph{brake} to stay at its current location. A car
will be removed once it reaches its destination at the edge of the grid.

Two cars \emph{collide} if their locations
overlap. A collision incurs a reward $r_{coll}=-10$, but does not
affect the simulation in any other way. To discourage a traffic jam,
each car gets reward of $\tau r_{time}=-0.01\tau$ at every time step,
where $\tau$ is the number time steps passed since the car arrived. Therefore, the total reward at time $t$ is:
\vspace{-2mm}
\[ r(t) = C^t r_{coll} + \sum_{i =1}^{N^t} \tau_i r_{time} ,\]
where $C^t$ is the number of collisions occurring at time $t$, and
$N^t$ is number of cars present. The simulation is terminated after 40
steps and is classified as a failure if one or more more
collisions have occurred. 

Each car is represented by one-hot binary vector set $\{n, l, r\}$,
that encodes its unique ID, current location and assigned route number
respectively. Each agent controlling a car can only observe other cars in
its vision range (a surrounding $3\times 3$ neighborhood), but it can communicate to
all other cars. The state vector $s_j$ for each
agent is thus a concatenation of all these vectors, having dimension
$3^2 \times |n| \times |l| \times |r|$.  

In \tab{traffic}(left), we show the probability of failure of a variety of
different model $\Phi$ and module $f$ pairs. Compared to the baseline models,
CommNet significantly reduces the failure rate for
all module types, achieving the best performance with LSTM module (a video showing this model before and after training can be found at \url{http://cims.nyu.edu/~sainbar/commnet}). 

We also explored how partial visibility within
the environment effects the advantage given by communication. As the
vision range of each agent decreases, the advantage of communication
increases as shown in \fig{junction}(right). 
Impressively, with zero visibility (the cars are driving
blind) the CommNet model is still able to succeed 90\% of the time. 

\tab{traffic}(right) shows the results on easy and hard versions of the game.
The easy version is a junction of two one-way roads, while 
the harder version consists from four connected junctions of two-way
roads. Details of the other game variations can be found in \append{traffic}. 
Discrete communication works well on the easy version, but the CommNet
with local connectivity gives the best performance on the hard case.

\begin{table}[b]
\center
\footnotesize
\setlength{\tabcolsep}{4pt}
\begin{tabular}{|l||c|c|c|} \hline
      & \multicolumn{3}{|c|}{Module $f()$ type} \\ 
Model $\Phi$              & MLP & RNN & LSTM \\ \hline
Independent               & \res{20.6}{14.1} & \res{19.5}{4.5} & \res{9.4}{5.6} \\ \hline
Fully-connected          & \res{12.5}{4.4} & \res{34.8}{19.7} & \res{4.8}{2.4} \\ \hline
Discrete comm.           & \res{15.8}{9.3} & \res{15.2}{2.1} & \res{8.4}{3.4}  \\ \hline \hline
CommNet         & \textbf{\res{2.2}{0.6}} & \textbf{\res{7.6}{1.4}} & \textbf{\res{1.6}{1.0}}  \\ \hline
\end{tabular}
\quad
\begin{tabular}{|l||c|c|c} \hline
      & \multicolumn{2}{|c|}{Other game versions} \\ 
Model $\Phi$              & Easy (MLP) & Hard (RNN) \\ \hline
Independent               & \res{15.8}{12.5} & \res{26.9}{6.0} \\ \hline
Discrete comm.          & \res{1.1}{2.4} & \res{28.2}{5.7} \\ \hline \hline
CommNet         & \textbf{\res{0.3}{0.1}}  & \res{22.5}{6.1}  \\ \hline
CommNet local        & -       & \textbf{\res{21.1}{3.4}} \\ \hline
\end{tabular}

\caption{Traffic junction task. Left: failure rates (\%) for
    different types of model and module function
    $f(.)$. CommNet consistently improves performance, over the
    baseline models. Right: Game variants. In the
    easy case, discrete communication does help, but still less than
    CommNet. On the
    hard version, local communication (see \secc{extra}) does at least
    as well as broadcasting to all agents.
}
\label{tab:traffic}
\vspace{-1mm}
\end{table}

\vspace{-2mm}
\subsubsection{Analysis of Communication}
\vspace{-2mm}
\begin{figure}[h!]
\centering
\includegraphics[width=4.5cm]{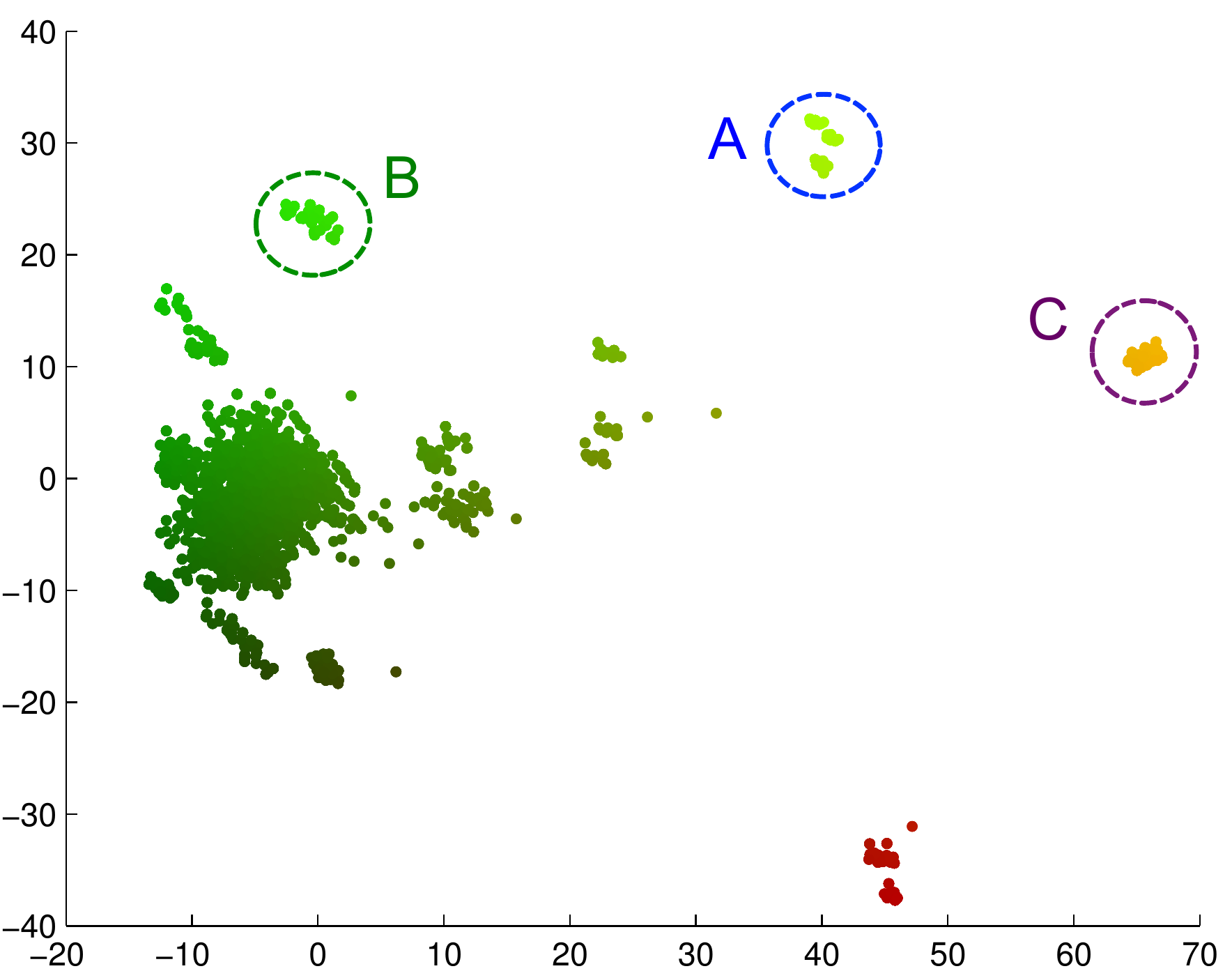}
\hspace{3mm}
\includegraphics[width=3.5cm]{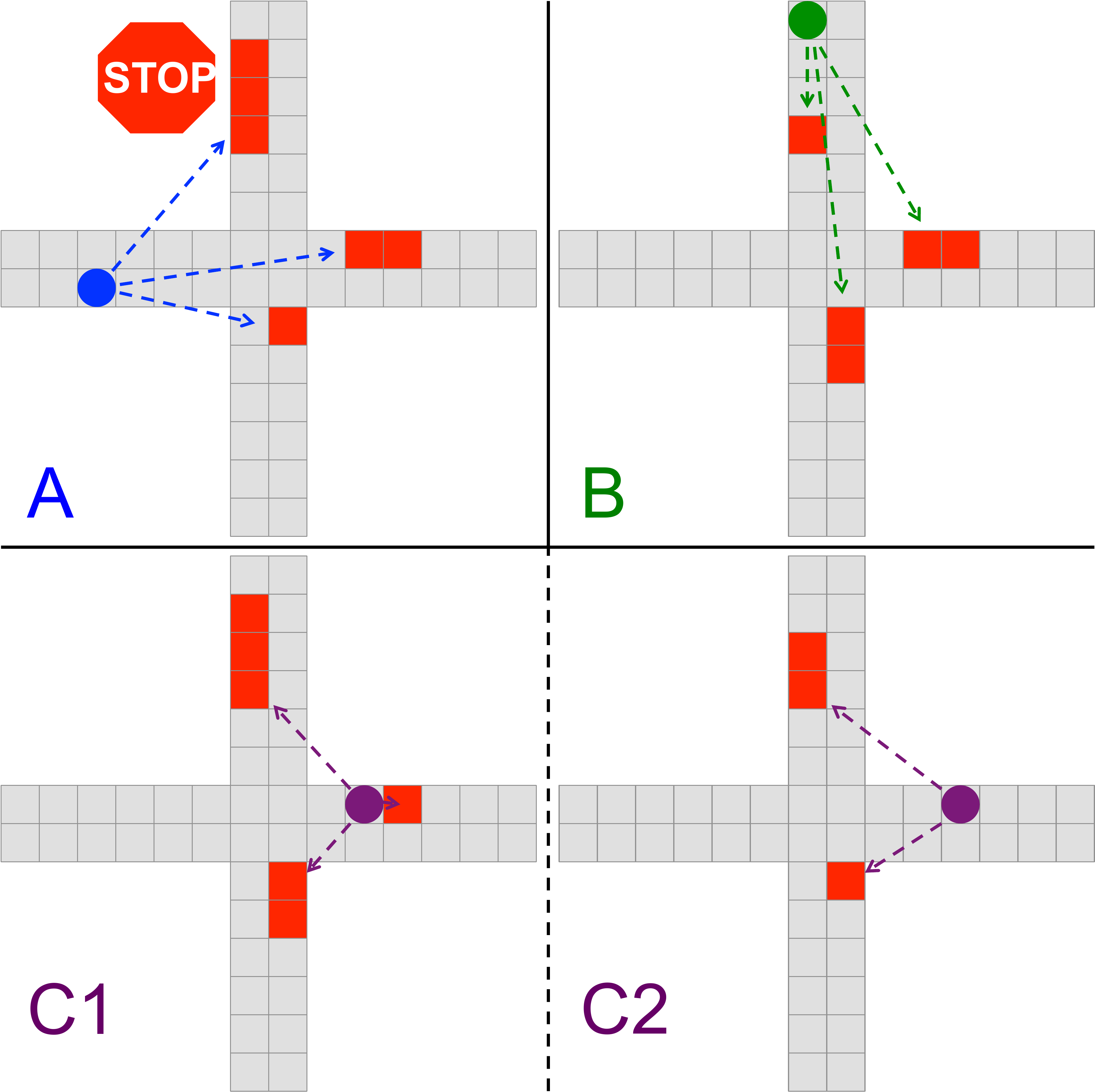}
\hspace{3mm}
\includegraphics[width=4.5cm]{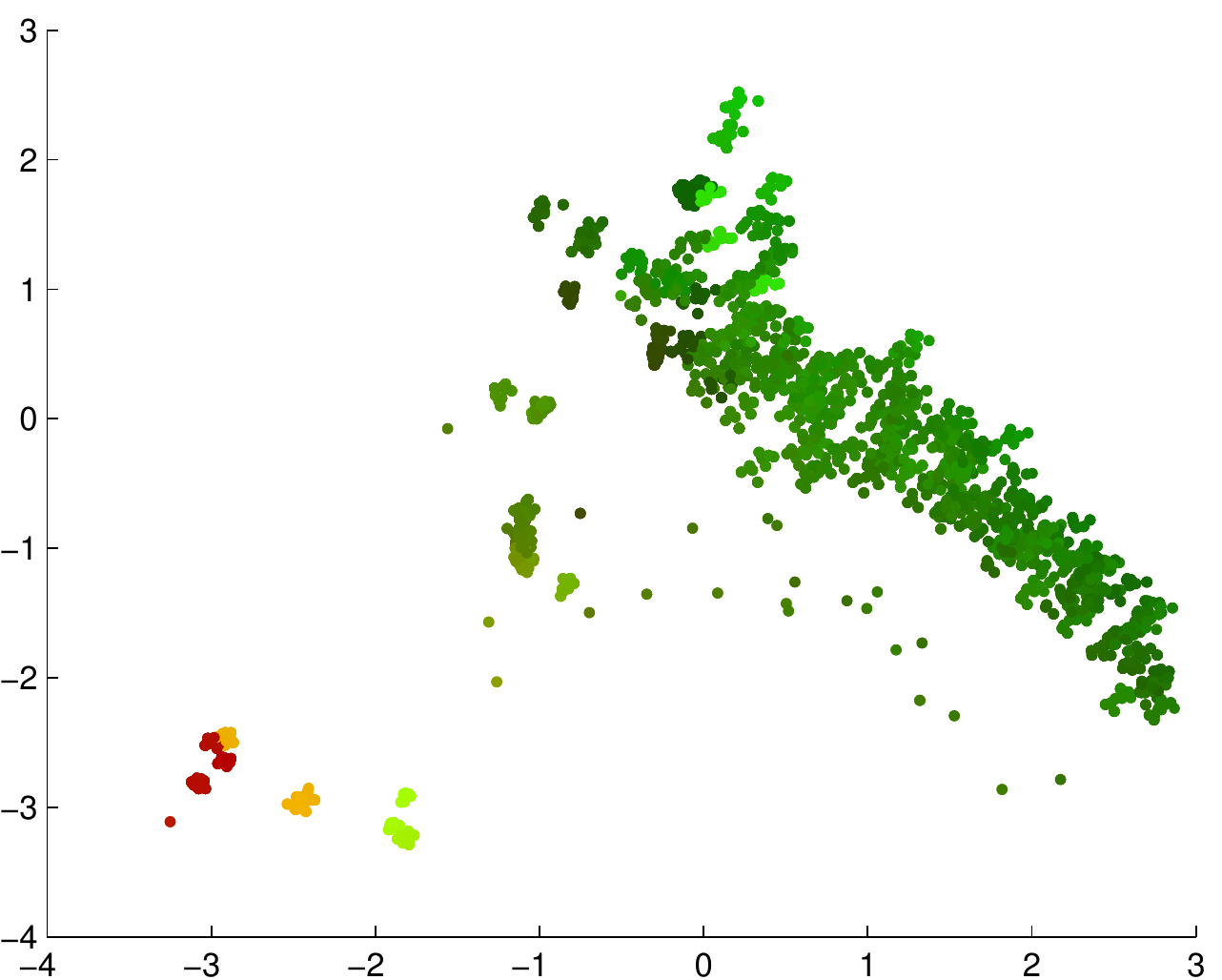}
\caption{Left: First two principal components of communication vectors
  $\tilde{c}$ from multiple runs on the traffic junction task
  \fig{junction}(left). While the majority are ``silent'' (i.e. have a
  small norm), distinct clusters are also present.  Middle: for three of these clusters,
  we probe the model to understand their
  meaning (see text for details). Right: First two principal
  components of hidden state vectors $h$ from the same runs as on the
  left, with corresponding color coding. Note how many of the
  ``silent'' communication vectors accompany non-zero hidden state
  vectors. This shows that the two pathways carry different
  information.
\vspace{-2mm} 
}
\label{fig:clusters}
\vspace{0mm}
\end{figure}

We now attempt to understand what the agents communicate when
performing the junction task. We start by recording the hidden state
$h^i_j$ of each agent and the corresponding \emph{communication vectors}
$\tilde{c}^{i+1}_j = C^{i+1}h^i_j$ (the contribution agent $j$ at step
$i+1$ makes to the hidden state of other agents). \fig{clusters}(left) and
\fig{clusters}(right) show the 2D PCA projections of the communication and hidden state
vectors respectively. These plots show a diverse range of hidden
states but far more clustered communication vectors, many of which are
close to zero. This suggests that while the hidden state carries
information, the agent often prefers not to communicate it to the
others unless necessary. This is a possible consequence of the broadcast
channel: if everyone talks at the same time, no-one can understand. 
See \append{analysis} for norm of communication vectors and brake locations.

To better understand the meaning behind the communication vectors, we
ran the simulation with only two cars and recorded their communication
vectors and locations whenever one of them braked. Vectors belonging
to the clusters A, B \& C in \fig{clusters}(left) were consistently
emitted when one of the cars was in a specific location, shown by the
colored circles in \fig{clusters}(middle) (or pair of locations for
cluster C). They also strongly correlated with the other car braking
at the locations indicated in red, which happen to be relevant to
avoiding collision.

\vspace{-3mm}
\subsubsection{Combat Task}
\vspace{-2mm}
We simulate a simple battle involving two opposing teams in a
$15\times15$ grid as shown in \fig{junction}(middle). Each team
consists of $m=5$ agents and their initial positions are sampled
uniformly in a $5\times 5$ square around the team center, which is
picked uniformly in the grid.  At each time step, an agent can perform
one of the following actions: move one cell in one of four directions;
attack another agent by specifying its ID $j$ (there are $m$ attack
actions, each corresponding to one enemy agent); or do nothing. If
agent A attacks agent B, then B's health point will be reduced by 1,
but only if B is inside the firing range of A (its surrounding
$3\times 3$ area). Agents need one time step of cooling down after an
attack, during which they cannot attack. All agents start with 3 health
points, and die when their health reaches 0.  A team will win if all
agents in the other team die.  The simulation ends when one team wins,
or neither of teams win within 40 time steps (a draw).

The model controls one team during training, and the other team
consist of bots that follow a hard-coded policy. The bot policy is to
attack the nearest enemy agent if it is within its firing range. If
not, it approaches the nearest visible enemy agent within visual range. An
agent is visible to all bots if it is inside the visual range of any
individual bot. This shared vision gives an advantage to the bot team.
When input to a model, each agent is represented by a set of one-hot
binary vectors $\{i, t, l, h, c\}$ encoding its unique ID, team ID,
location, health points and cooldown. A model controlling an agent
also sees other agents in its visual range ($3 \times 3$ surrounding
area). The model gets reward of -1 if the team loses or draws at the
end of the game. In addition, it also get reward of $-0.1$ times the
total health points of the enemy team, which encourages it to attack
enemy bots.

\begin{table}[h]
\center
\footnotesize
\setlength{\tabcolsep}{3pt}
\begin{tabular}{|l|c|c|c|} \hline
              & \multicolumn{3}{|c|}{Module $f()$ type} \\ 
Model $\Phi$             & MLP & RNN & LSTM \\ \hline
Independent               & \res{34.2}{1.3} & \res{37.3}{4.6} & \res{44.3}{0.4}  \\ \hline
Fully-connected          & \res{17.7}{7.1} & \res{2.9}{1.8} & \res{19.6}{4.2}  \\ \hline
Discrete comm.           & \res{29.1}{6.7} & \res{33.4}{9.4} & \res{46.4}{0.7} \\ \hline \hline
CommNet         & \textbf{\res{44.5}{13.4}} & \textbf{\res{44.4}{11.9}} & \textbf{\res{49.5}{12.6}} \\ \hline
\end{tabular}
\:
\begin{tabular}{|l|c|c|c|} \hline
              & \multicolumn{3}{|c|}{Other game variations (MLP)} \\ 
Model $\Phi$             & $m=3$ & $m=10$ & $5\times 5$ vision \\ \hline
Independent               &\res{29.2}{5.9} & \res{30.5}{8.7} & \res{60.5}{2.1} \\ \hline
CommNet         & \textbf{\res{51.0}{14.1}} & \textbf{\res{45.4}{12.4}} & \textbf{\res{73.0}{0.7}} \\ \hline
\end{tabular}

\caption{Win rates (\%) on the combat task for different communication
approaches and module choices. Continuous consistently outperforms the
other approaches. The fully-connected baseline does worse than the independent model without 
communication. On the right we explore the effect of varying
the number of agents $m$ and agent visibility. Even with 10 agents on
each team, communication clearly helps.}
\label{tab:combat}
\end{table}

\tab{combat} shows the win rate of different module choices with various types of
model. Among different modules, the LSTM achieved the best
performance. Continuous communication with CommNet improved all module
types. Relative to the independent controller, the fully-connected model degraded performance,
but the discrete communication improved LSTM module type.
We also explored several variations of the task: varying the number of
agents in each team by setting $m=3,10$, and increasing visual range
of agents to $5\times 5$ area. The result on those tasks are shown on
the right side of \tab{combat}. Using CommNet model consistently 
improves the win rate, even with the
greater environment observability of the 5$\times$5 vision case.

\vspace{-3mm}
\subsection{bAbI Tasks}
\vspace{-3mm}
\label{sec:babi}
We apply our model to the bAbI~\cite{Weston15} toy Q \& A dataset, which
consists of 20 tasks each requiring different kind of reasoning. The goal
is to answer a question after reading a short story. We can
formulate this as a multi-agent task by giving each sentence of the
story its own agent. Communication among agents allows them to
exchange useful information necessary to answer the question.

The input is $\{s_1, s_2, ..., s_J, q\}$, where $s_j$ is $j$'th
sentence of the story, and $q$ is the question sentence. We use the same
encoder representation as \cite{end2endmemnn} to convert them to
vectors. The $f(.)$ module consists of a two-layer MLP with ReLU
non-linearities. After $K=2$ communication steps, we add the final 
hidden states together and pass it through a softmax decoder layer
to sample an output word $y$. The model is trained in a
supervised fashion using a cross-entropy loss between $y$ and the
correct answer $y^*$. The hidden layer size is set to 100 and weights are
initialized from $N(0,0.2)$. We train the model for 100 epochs with
learning rate 0.003 and mini-batch size 32 with Adam
optimizer~\cite{kingma2015} ($\beta_1=0.9, \beta_2=0.99,
\epsilon=10^{-6}$). We used 10\% of training data as validation set to
find optimal hyper-parameters for the model.

Results on the 10K version of the bAbI task are shown in \tab{babi},
along with other baselines (see \append{babi} for a detailed
breakdown). Our model outperforms the LSTM baseline, but is worse than
the MemN2N model \cite{end2endmemnn}, which is specifically designed
to solve reasoning over long stories. However, it successfully solves
most of the tasks, including ones that require information sharing
between two or more agents through communication.

\begin{table}[h]
\centering
\small
\begin{tabular}{|l|c|c|} 
\hline
                  & Mean error (\%) & Failed tasks (err. > 5\%) \\ \hline
LSTM~\cite{end2endmemnn}        & 36.4 & 16 \\ \hline
MemN2N~\cite{end2endmemnn}      & 4.2 &  3 \\ \hline
DMN+~\cite{XiongMS16}           & \textbf{2.8} &  \textbf{1} \\ \hline
Independent (MLP module)  & 15.2 &  9 \\ \hline
\hline
CommNet (MLP module)  & 7.1 &  3 \\ \hline
\end{tabular}
\caption{Experimental results on bAbI tasks.
}
\label{tab:babi}
\vspace{-3mm}
\end{table}

\vspace{-4mm}
\section{Discussion and Future Work}
\vspace{-2mm} 
We have introduced CommNet, a simple controller for MARL that is able
to learn continuous communication between a dynamically changing set of agents. Evaluations on four
diverse tasks clearly show the model outperforms models without
communication, fully-connected models, and models using discrete communication.  Despite the simplicity of the
broadcast channel, examination of the traffic task reveals the model
to have learned a sparse communication protocol that conveys
meaningful information between agents. Code for our model (and baselines) can be found at \url{http://cims.nyu.edu/~sainbar/commnet/}.

 One aspect of our model that we did not fully
exploit is its ability to handle heterogenous agent types and we hope
to explore this in future work.  Furthermore, we believe the model will scale gracefully to large numbers of agents, perhaps requiring more sophisticated connectivity structures; we also leave this to future work.

\vspace{-2mm}
\subsection*{Acknowledgements}
\vspace{-2mm} 
The authors wish to thank Daniel Lee and Y-Lan Boureau for their
advice and guidance. Rob Fergus is grateful for the support of CIFAR.

\small
\vspace{-3mm}
\bibliography{bibliography}
\bibliographystyle{ieee}

\appendix

\section{Reinforcement Training}
\label{app:reinforce}
We use policy gradient \cite{Williams92simplestatistical} with a state 
specific baseline for delivering a gradient to the model.  Denote the
states in an episode by $s(1),...,s(T)$, and the actions taken at each
of those states as $a(1),...,a(T)$, where $T$ is the length of the
episode. The baseline is a scalar function of the states
$b(s,\theta)$, computed via an extra head on the model producing the
action probabilities. Beside maximizing the expected reward with
policy gradient, the models are also trained to minimize the distance
between the baseline value and actual reward. Thus after finishing
an episode, we update the model parameters $\theta$ by
\begin{equation}
\Delta \theta = \sum_{t=1}^T \left[ 
  \frac{\partial \log p(a(t)| s(t), \theta)}{\partial \theta} 
  \left(\sum_{i=t}^T r(i) - b(s(t), \theta) \right) 
   - \alpha
  \frac{\partial}{\partial \theta} 
  \left(\sum_{i=t}^T r(i) - b(s(t), \theta) \right)^2 
\right].
\label{eq:RL}
\end{equation}
Here $r(t)$ is reward given at time $t$, and the hyperparameter $\alpha$ is for balancing the reward and the baseline objectives, which set to 0.03 in all experiments.

\section{Lever Pulling Task Analysis}
\label{app:lever}
\begin{figure}[h!]
\centering
\includegraphics[width=8cm,trim=4cm 4cm 4cm 4cm,clip=true]{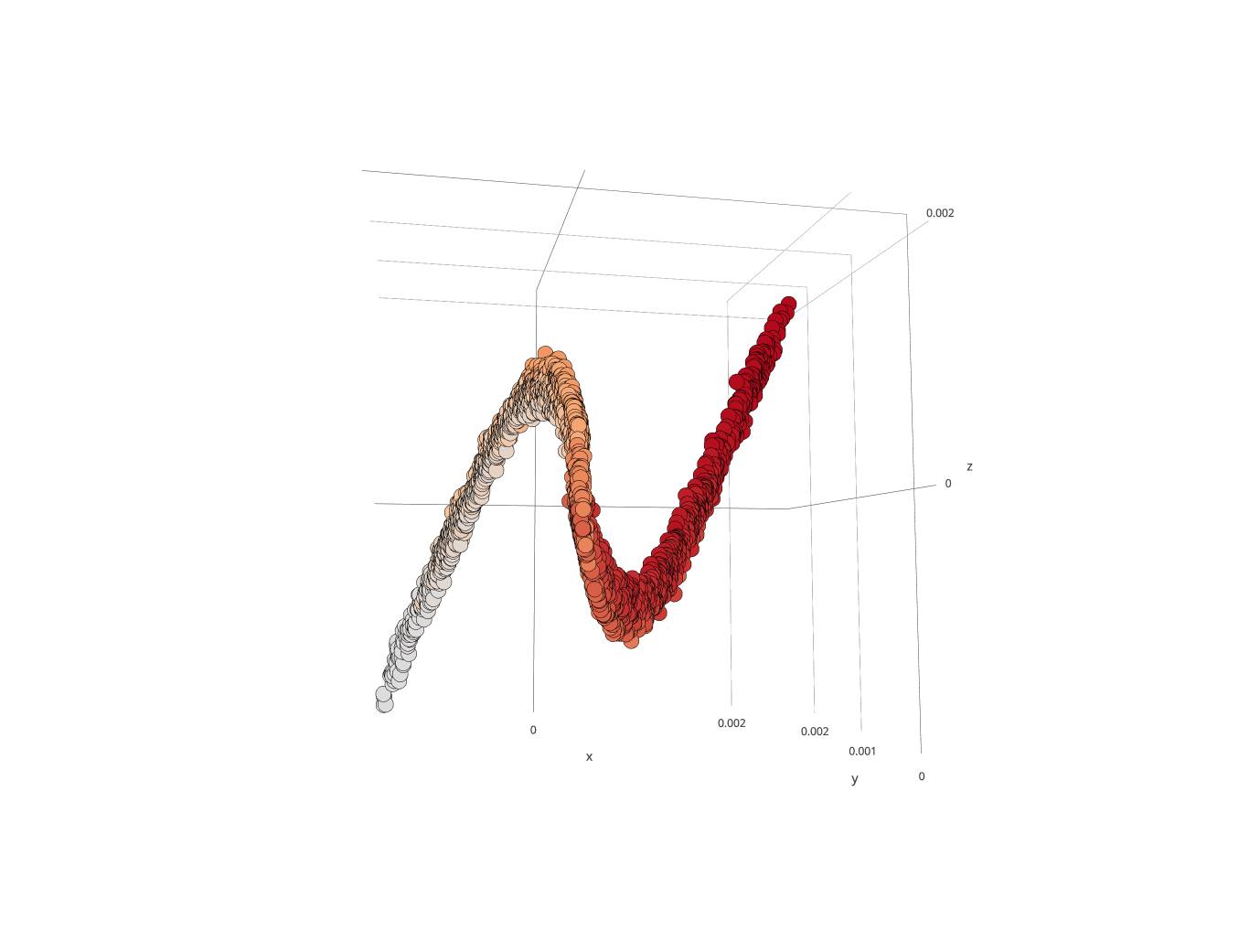}
\caption{3D PCA plot of hidden states of agents}
\label{fig:leverpca}
\end{figure}

Here we analyze a CommNet model trained with supervision on the lever pulling task. 
The supervision uses the sorted ordering of agent IDs to assign target actions.
For each agent, we concatenate its hidden layer activations during game playing.
\fig{leverpca} shows 3D PCA plot of those vectors, where color intensity represents agent's ID.
The smooth ordering suggests that agents are communicating their IDs, enabling them to solve the task.

\section{Details of Traffic Junction}
\label{app:traffic}
We use curriculum
learning~\cite{Bengio09} to make the training easier. In first 100
epochs of training, we set $p_\text{arrive}=0.05$, but linearly
increased it to $0.2$ during next 100 epochs. Finally, training
continues for another 100 epochs. The learning rate is fixed at 0.003 throughout.
We also implemented additional easy and hard versions of the game,
the latter being shown in Fig.2. 

The easy version is a
junction of two one-way roads on a $7\times7$ grid. There are two
arrival points, each with two possible routes. During curriculum, we
increase $N_\text{total}$ from 3 to 5, and $p_\text{arrive}$ from 0.1
to 0.3.

\begin{figure}[h!]
\centering
\includegraphics[width=0.3\linewidth]{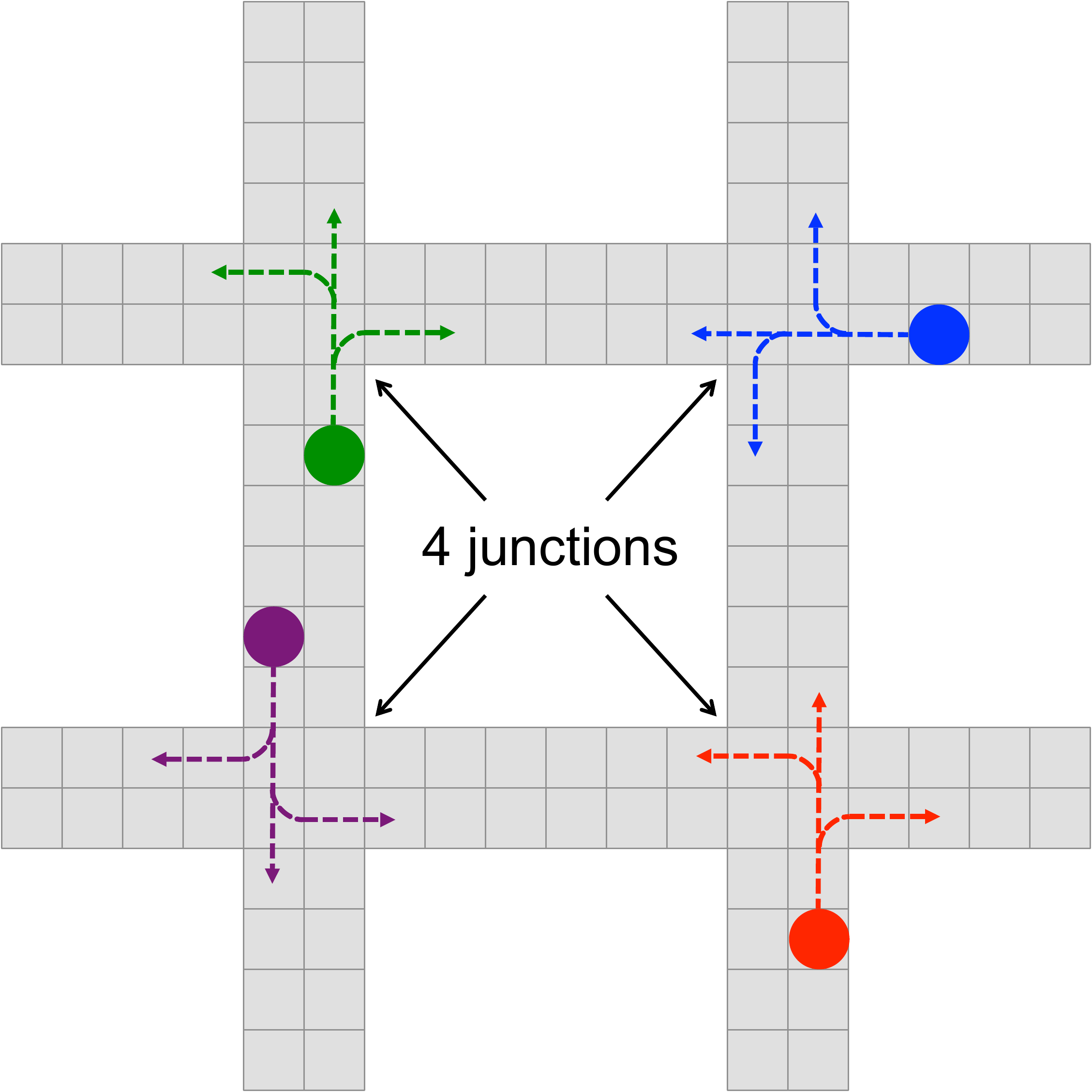}
\caption{A harder version of traffic task with four connected junctions. }
\label{fig:junction4}
\end{figure}

The harder version consists from four connected junctions of two-way
roads in $18\times18$ as shown in \fig{junction4}. There are 8
arrival points and 7 different routes for each arrival point. We set
$N_\text{total}=20$, and increased $p_\text{arrive}$ from 0.02 to 0.05
during curriculum.

\section{Traffic Junction Analysis}
\label{app:analysis}
\begin{figure}[h!]
\centering
\includegraphics{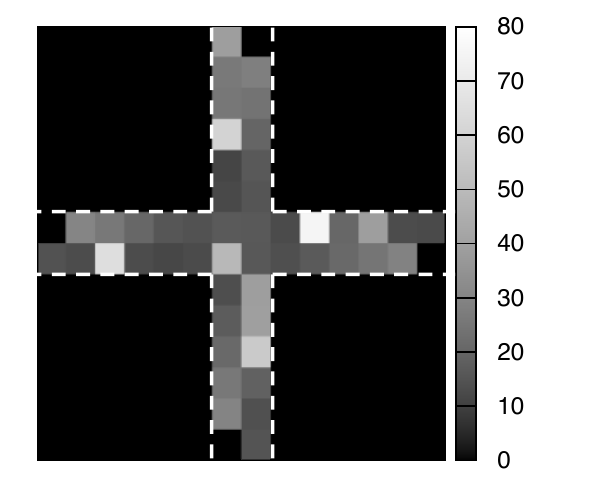}
\hspace{5mm}
\includegraphics{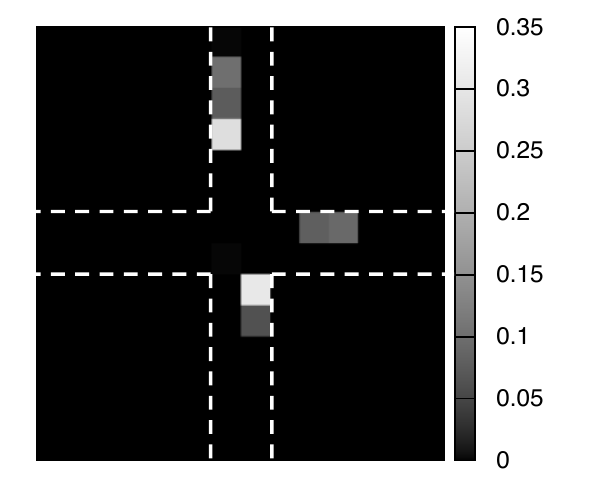}
\caption{(left) Average norm of communication vectors (right) Brake locations}
\label{fig:norms}
\end{figure}

Here we visualize the average norm of the communication vectors in \fig{norms}(left)
and brake locations over the $14\times 14$ spatial grid in \fig{norms}(right). 
In each of the four incoming directions, there is one location where communication signal is stronger.
The brake pattern shows that cars coming from left never yield to other directions. 

\section{bAbI Tasks Details}
\label{app:babi}

Here we give further details of the model setup and training, as well
as a breakdown of results in \tab{babi2}.

Let the task be $\{s_1, s_2, ..., s_J, q, y^*\}$, where $s_j$ is $j$'th sentence of story, $q$ is the question sentence and $y^*$ is the correct answer word (when answer is multiple words, we simply concatenate them into single word). Then the input to the model is
\[ h_j^0 = r(s_j, \theta_0),\quad c_j^0 = r(q, \theta_q) .\]
Here, we use simple position encoding~\cite{end2endmemnn} as $r$ to convert sentences into fixed size vectors. 
Also, the initial communication is used to broadcast the question to all agents. Since the temporal ordering of sentences is relevant in some tasks, we add special temporal word ``$t=J-j$'' to $s_j$ for all $j$.

For $f$ module, we use a 2 layer network with skip connection, that is 
\[ h_j^{i+1} = \sigma(W_i \sigma(H^i h_j^i + C^i c_j^i + h_j^0 ) ), \]
where $\sigma$ is ReLU non-linearity (bias terms are omitted for clarity). After $K=2$ communication steps, the model outputs an answer word by 
\[ y = Softmax(D \sum_{j=1}^J h_j^K) \]
Since we have the correct answer during training, we will do
supervised learning by using cross entropy cost on $\{y^*, y\}$. The
hidden layer size is set 100 and weights are initialized from
$N(0,0.2)$. We train the model 100 epochs with learning rate 0.003 and
mini-batch size 32 with Adam optimizer~\cite{kingma2015}
($\beta_1=0.9, \beta_2=0.99, \epsilon=10^{-6}$). We used 10\% of
training data as validation set to find optimal hyper-parameters for
the model.

\begin{table}[h]
\centering
\small
\setlength{\tabcolsep}{4pt}
\begin{tabular}{|l|c|c|c|c|c|c|c|c|c|} 
\hline
                  & \multicolumn{7}{|c|}{Error on tasks (\%)} & Mean error & Failed tasks \\ \cline{2-8}
                  & 2 & 3 & 15 & 16 & 17 & 18 & 19 & (\%) & (err. > 5\%) \\ \hline
LSTM~\cite{end2endmemnn}        & 81.9 & 83.1 & 78.7 & 51.9 & 50.1 & 6.8 & 90.3 & 36.4 & 16 \\ \hline
MemN2N~\cite{end2endmemnn}      & \textbf{0.3} & 2.1 & \textbf{0.0} & 51.8 & 18.6 & 5.3 & 2.3 &  4.2 &  3 \\ \hline
DMN+~\cite{XiongMS16}           & \textbf{0.3} & \textbf{1.1} & \textbf{0.0} & \textbf{45.3} & 4.2 & 2.1 & \textbf{0.0} & \textbf{2.8} &  \textbf{1} \\ \hline
Neural Reasoner+~\cite{Peng15}          & - & - & - & - & \textbf{0.9} & - & 1.6 & - &  - \\ \hline
Independent (MLP module)  & 69.0 & 69.5 & 29.4 & 47.4 & 4.0 & \textbf{0.6} & 45.8 & 15.2 &  9 \\ \hline
\hline
CommNet (MLP module)  & 3.2 & 68.3 & \textbf{0.0} & 51.3 & 15.1 & 1.4 & \textbf{0.0} & 7.1 &  3 \\ \hline
\end{tabular}
\caption{Experimental results on bAbI tasks. Only showing some of the task with high errors.}
\label{tab:babi2}
\end{table}

\end{document}